\documentclass{ecai}

\usepackage{iftex}
\ifPDFTeX
\fi

\usepackage[T1]{fontenc}
\usepackage{enumitem}
\usepackage{tabularx}
\usepackage{graphicx}
\usepackage{mdframed}
\usepackage{xcolor}
\usepackage{tcolorbox}
\usepackage{makecell}
\usepackage{listings}
\usepackage{array}     
\usepackage{caption}   
\usepackage{subcaption}
\usepackage{tikz}  
\usepackage{lipsum}  
\usepackage{booktabs}
\usepackage{multirow}
\usepackage{colortbl}
\usepackage{xspace}
\usepackage{pifont}
\usepackage{threeparttable}
\usepackage{adjustbox}
\usepackage{tikz}
\usepackage{pgf}
\usepackage{tcolorbox}
\usepackage{multicol}
\usepackage{hyperref}
\usepackage{tabularx}
\usepackage{longtable}
\usepackage{placeins}
\usepackage{amsmath}
\usepackage{amssymb}

\newcommand{\sysname}  {\textsc{DoTabler}\xspace}

\makeatletter
\newcommand{\silentfootnote}[1]{%
  \begingroup
    \addtocounter{footnote}{0}%
    \let\@makefnmark\relax 
    \let\@makefntext\@firstofone
    \footnotetext[\value{footnote}]{#1}%
  \endgroup
}
\makeatother

\newcommand{\BibTeX}{B\kern-.05em{\sc i\kern-.025em b}\kern-.08em\TeX}

\definecolor{myblue}{RGB}{70,130,180}
\definecolor{darkred}{RGB}{139, 0, 0}

\definecolor{commentgreen}{RGB}{2,112,10}
\definecolor{eminence}{RGB}{108,48,130}
\definecolor{weborange}{RGB}{255,165,0}
\definecolor{frenchplum}{RGB}{129,20,83}

\definecolor{added}{rgb}{0.9, 1.0, 0.9}
\definecolor{removed}{rgb}{1.0, 0.9, 0.9}
\definecolor{codebg}{rgb}{0.95, 0.95, 0.95}

\begin{document}

\begin{frontmatter}

\paperid{8463}

\title{From Surface to Semantics: Semantic Structure Parsing for Table-Centric Document Analysis}

\author{
\begin{center}
\textbf{Xuan Li},
\textbf{Jialiang Dong\textsuperscript{*}},
\textbf{Raymond Wong\textsuperscript{*}}
\end{center}
}

\address{University of New South Wales, Sydney, Australia}

\begin{abstract}
Documents are core carriers of information and knowledge, 
    with broad applications in finance, 
    healthcare, 
    and scientific research.
Tables, 
    as the main medium for structured data, 
    encapsulate key information and are among the most critical document components.
Existing studies largely focus on surface-level tasks such as layout analysis, 
    table detection, 
    and data extraction, 
    lacking deep semantic parsing of tables and their contextual associations. 
This limits advanced tasks like cross-paragraph data interpretation and context-consistent analysis. 
To address this, 
    we propose \sysname,
    a table-centric semantic document parsing framework designed to uncover deep semantic links between tables and their context. 
\sysname leverages a custom dataset and domain-specific fine-tuning of pretrained models, 
    integrating a complete parsing pipeline to identify context segments semantically tied to tables.
Built on this semantic understanding, 
    \sysname implements two core functionalities: 
    table-centric document structure parsing and domain-specific table retrieval,
    delivering comprehensive table-anchored semantic analysis and precise extraction of semantically relevant tables. 
Evaluated on nearly 4,000 pages with over 1,000 tables from real-world PDFs, 
    \sysname achieves over 90\% Precision and F1 scores, 
    demonstrating superior performance in table-context semantic analysis and deep document parsing compared to advanced models such as GPT-4o.
\end{abstract}

\end{frontmatter}

\silentfootnote{\textsuperscript{*}~Corresponding authors. Emails: jialiang.dong@unsw.edu.au, ray.wong@unsw.edu.au}

\section{Introduction}
Documents are vital carriers of information across domains such as government, enterprise, and science, playing a foundational role in sectors like finance, healthcare, and academia~\cite{Li2024FinancialExtraction, hein2025prompts, li2020tablebank, zhong2019publaynet}. As noted by UNESCO, they are essential for global knowledge transmission and cultural preservation~\cite{unesco_memory}. Among document components, tables are the primary medium for structured data, often central to industrial document analysis tasks. For example, in the financial sector, analysts often need to retrieve revenue definitions along with relevant tables from reports. In the legal domain, contract reviewers must connect clauses with compensation tables. In the public sector, policy analysts frequently extract demographic summaries and related statistics from lengthy government reports. These tasks involve heterogeneous document structures, which makes the semantic table-text association essential for efficient and accurate information access.
\begin{figure}[t]
    \centering
    \begin{subfigure}{0.9\linewidth}
        \centering
        \includegraphics[width=\linewidth]{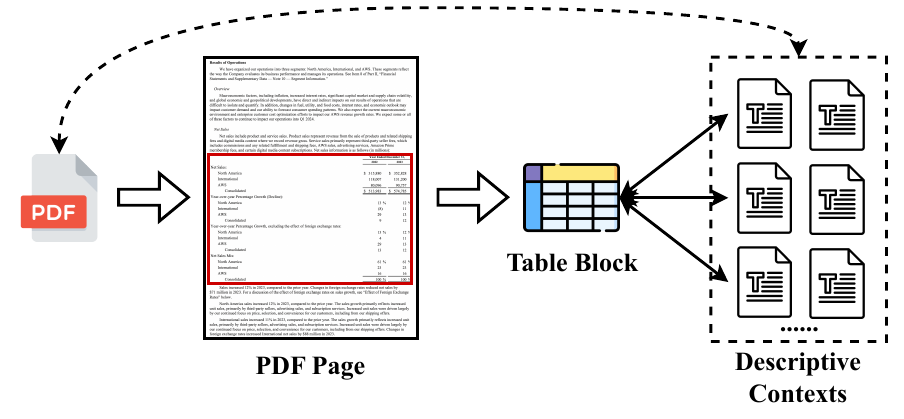}
        \caption{Instance of Table-Centric Parsing}
        \vspace{15pt}
        \label{subfig:instance_a}
    \end{subfigure}
    \vspace{10pt}
    \begin{subfigure}{0.9\linewidth}
        \centering
        \includegraphics[width=\linewidth]{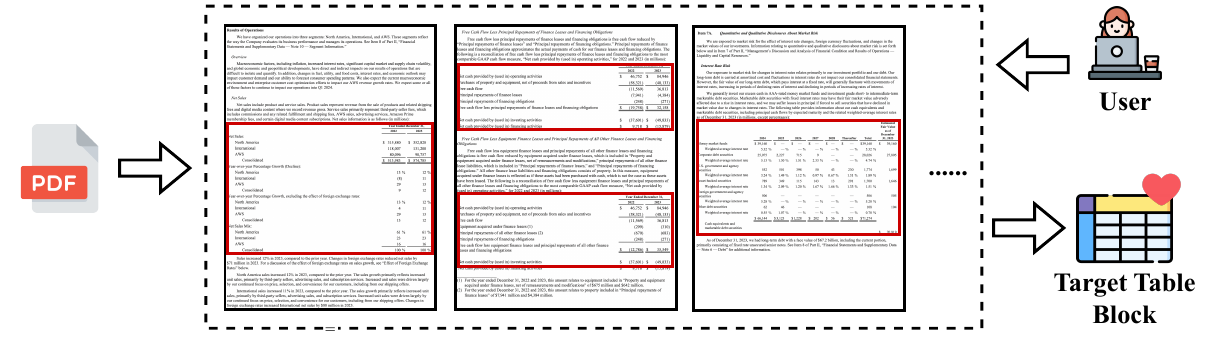}
        \vspace{1pt}
        \caption{Instance of Domain-Specific Table Retrieval}
        \label{subfig:instance_b}
    \end{subfigure}
    \caption{Example Applications of Semantic Structure Analysis}
    \vspace{15pt}
    \label{fig:parsing_time_overhead}
\end{figure}

Extensive research has explored automated analysis and extraction of information from documents and their tables \cite{ wan2024omniparser}.
As the dominant document format, 
    PDF is widely adopted due to its lightweight nature, 
    cross-platform compatibility, 
    and consistent layout \cite{zhong2019publaynet}. 
However, 
    its page-based architecture encodes text and tables as embedded graphical elements, 
    making direct parsing difficult and significantly increasing analysis complexity~\cite{Li2024FinancialExtraction}. 
Current studies mainly focus on shallow visual-level structural analysis,
    broadly falling into three directions.
(1) document layout analysis~\cite{gu2022xylayoutlm, huang2022layoutlmv3}, 
    which identifies regions such as text blocks, tables, and figures; 
(2) table detection~\cite{zhao2024tabpedia,zhong2019publaynet}, 
    which localizes tables within pages;
and (3) table structure extraction and recognition~\cite{nassar2022tableformer}, 
    which reconstructs tables by parsing their structure and recognizing embedded text and data. 
Despite these advances, 
    existing methods largely remain at the visual structural level, 
    converting embedded content into machine-readable forms but lacking a deep understanding of document semantics, table content, and contextual relationships \cite{zhang2020document, huang2022layoutlmv3}.

For instance,
    as depicted in Figure~\ref{subfig:instance_a}, 
    for table data, 
    merely extracting content and converting it to machine-readable formats is insufficient; 
    analyzing contextual information is equally crucial. 
This context explains the data, 
    its intended use, 
    and underlying logical relationship
    foundations for semantic understanding and advanced reasoning. 
Besides, 
    as shown in Figure~\ref{subfig:instance_b}, 
    documents often contain multiple tables, 
    but practical analysis typically focuses on task-relevant ones. 
In large document collections,
    efficiently retrieving domain-specific tables is key to effective information use.
Thus, 
    shallow visual analysis is inadequate for complex tasks, 
    deep semantic parsing is indispensable for robust information extraction.

However, 
    implementing table-centric semantic document and table parsing presents multiple challenges.
First, 
    as a page-description format, 
    PDF embeds content as images or vector graphics without inherent structural annotations, 
    making elements such as text and tables difficult to parse directly \cite{wan2024omniparser}.
Currently, 
    no comprehensive solutions exist for document-level semantic segmentation and extraction, 
    and achieving efficient and accurate semantic partitioning in complex documents remains a significant challenge.
Second,
    analyzing the semantic relationship between table blocks and text blocks constitutes another core challenge.
Both are unstructured, 
    lacking explicit links,
    which hinders direct semantic association. 
While natural language processing (NLP) techniques can extract implicit semantic relations, 
    the absence of high-quality datasets modeling table-context associations in documents limits the training of traditional NLP models \cite{ma2024mmlongbench, tang2023unifying}. 
Moreover, 
    although large language models (LLMs) possess strong general understanding capabilities, 
    their performance is constrained by training data, 
    and pervasive hallucination issues further impede precise semantic relation modeling \cite{ji2023survey}.

To this end, 
    we propose \sysname, 
    which to the best of our knowledge is the first framework for table-centric semantic document parsing. 
\sysname integrates multiple shallow-level document analysis modules to construct a complete preprocessing pipeline,
    including document segmentation, 
    layout analysis, 
    and optical character recognition (OCR), 
    providing support for subsequent semantic parsing. 
Based on this, 
    we developed the first semantic-level dataset modeling table-text relationships and trained the Table-Text Association Model (TTAM) as the core component.
Leveraging TTAM, 
    \sysname implements two key functionalities:
    document semantic structure parsing and domain-specific table retrieval.
We evaluated \sysname on nearly 4,000 pages of real-world PDF documents containing over 1,000 tables. 
The results show that \sysname achieved the precision and F1 scores of over 90\% in the semantic analysis of the table context, significantly outperforming advanced models such as GPT-4o, Gemini-2.0, and Claude-3.5, while delivering orders of magnitude improvements in execution efficiency.
In summary, the key contributions of this study are as follows:

\begin{itemize}
    \item We propose the first PDF semantic-level dataset modeled around table-centric structures and train the TTAM to effectively analyze relationships between tables and their contextual content.

    \item We design \sysname, which integrates a complete document preprocessing pipeline and semantic relationship analysis of PDF elements, enabling both semantic structure parsing and domain-specific table retrieval.

    \item We conduct a comprehensive evaluation of \sysname on nearly 4,000 pages of real-world PDF documents, demonstrating its superior performance and practical utility in semantic structure parsing. The source code of \sysname and the experiment datasets are available at \href{https://github.com/xuan084/DoTabler2025}{https://github.com/xuan084/DoTabler2025}.
\end{itemize}

\section{Related Works}
\subsection{Table Extraction and Recognition}

Modern Table Extraction (TE) frameworks often adapt generic object detection models,
    such as Faster R-CNN and Mask R-CNN,
    to the specific tasks of table detection and segmentation, achieving substantial performance improvements~\cite{li2020tablebank}. 
More advanced models, 
    including Cascade Mask R-CNN \cite{prasad2020cascadetabnet} and Transformer-based DETR \cite{smock2022pubtables}, have further enhanced detection precision, particularly for complex layouts. Enhancements like Deformable DETR (DDETR) improve multi-scale feature representation, mitigating the convergence issues and performance limitations of standard DETR models \cite{zhu2020deformable}. In addition, Tc-OCR, a hybrid framework that integrates DETR, Cascade TabNet, and PP OCR v2 into a hybrid architecture to improve table extraction in scanned and noisy PDFs \cite{anand2023tc}. Building on this, retrieval-augmented OCR models trained on domain-specific datasets yield improved text recognition for tables in financial, legal, and regulatory documents \cite{saleh5134287enhancing}. However, relatively few studies have explored context-aware information extraction that incorporates both tables and their surrounding textual context. 

\subsection{Key Information Extraction}

Key Information Extraction (KIE) from documents centers on accurately 
    identifying and structuring semantically meaningful textual content. Existing KIE approaches can be broadly categorized into OCR-dependent and OCR-free models~\cite{wan2024omniparser}.

The OCR-dependent methods traditionally rely on sequence labeling of OCR 
    outputs, often enhanced by layout-aware or graph-based representations that capture spatial and structural relationships between text segments~\cite{huang2022layoutlmv3, gu2022xylayoutlm, appalaraju2021docformer}. Auxiliary detection and linking models are introduced to model complex interdependencies among text blocks~\cite{yang2023modeling, yu2023structextv2}. Recent generation-based approaches frame KIE as a structured generation problem, simplifying decoding by directly generating entities as key-value pairs, further improving adaptability across tasks \cite{tang2023unifying}. In contrast, OCR-free methods aim to bypass traditional OCR pipelines 
    entirely by incorporating text-reading capabilities directly into end-to-end architectures. Models like Donut and other sequence-to-sequence (Seq2Seq) frameworks \cite{cao2023attention, kim2022ocr} are pre-trained with document image-to-text generation objectives and can directly produce structured text representations. 

\subsection{LLM-based Text Semantic Analysis}
The advent of LLMs introduces the ability to model rich contextual embeddings, enabling a more nuanced understanding of the semantic relationships between entities \cite{zhang2020document}. Models like DocuNet and Longformer incorporate sparse attention or sliding window mechanisms, allowing for effective modeling of documents with thousands of tokens \cite {zhang2021document, beltagy2020longformer}. The encoder-decoder architectures such as GPT-3, T5, and Flan-T5 support multirelation and multihop extraction without relying on rigid predefined relation types~\cite{mastropaolo2021studying, wan2023gpt,wadhwa2023revisiting}. 

For the extraction of table content relations, 
    LLMs have been successfully applied for the extraction of clinical information \cite{hein2025prompts}. 
Beyond predefined table structures, 
    DynoClass, 
    a self-adaptive system,
    detects table classes dynamically without requiring predefined ontologies. 
This approach is particularly beneficial for evolving datasets, 
    such as those encountered in business intelligence and market research,
    where table formats frequently change \cite{wang2024dynoclass}.

Despite these advances, 
    challenges remain in integrating multimodal document signals (e.g., text + tables) and ensuring that extracted relations are coherent and factually consistent across modalities. 
Our work builds on this line of research by combining document understanding with specialized modules for table extraction, 
    enabling a unified approach that captures both text-based and table-based semantic relationships.

\section{Methodology}
\begin{figure*}[t]
    \centering
    \includegraphics[width=\textwidth]{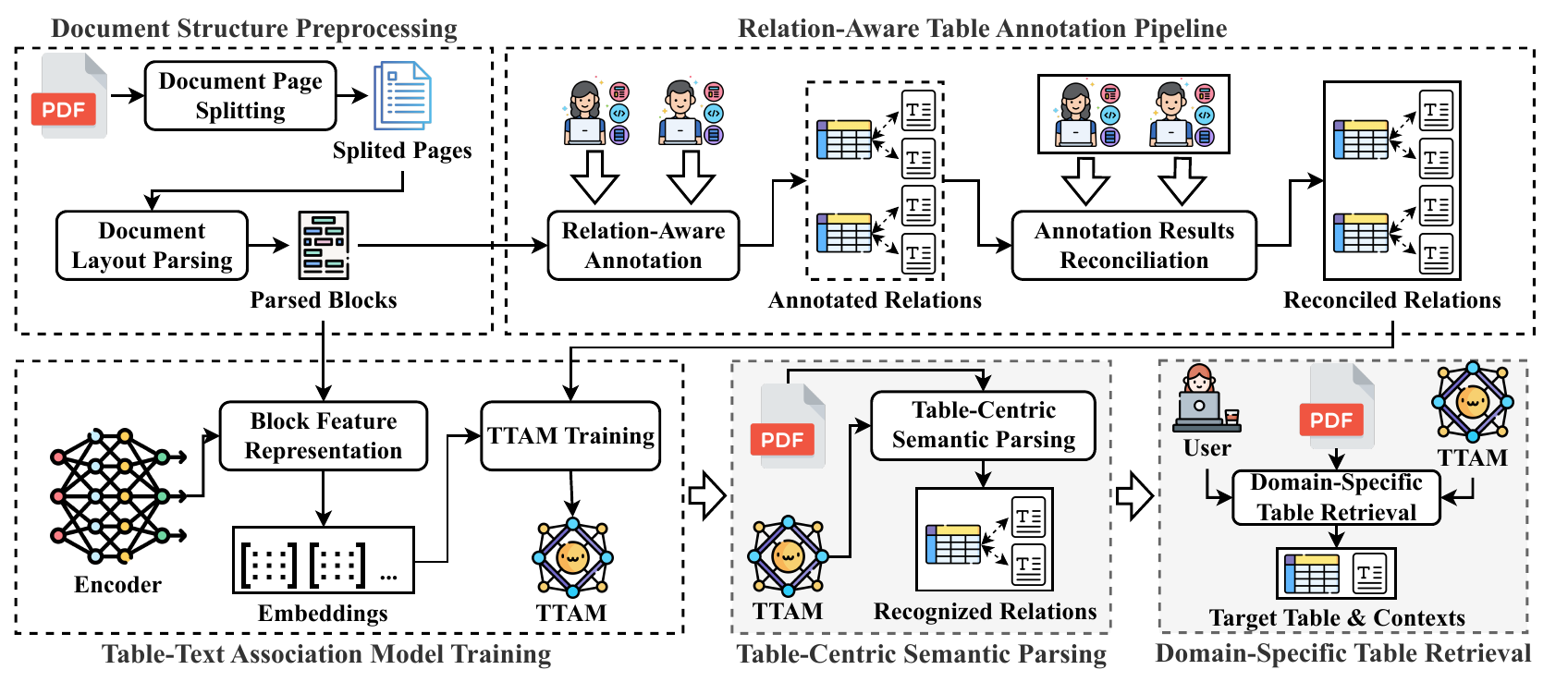}
    \caption{The overall workflow of \sysname}
    \label{fig:overview}
    \vspace{5pt}
\end{figure*}

The overall workflow of \sysname is illustrated in Figure~\ref{fig:overview}.
Given a target PDF, 
    we begin with Document Structure Preprocessing~(\ref{subsec:document_preprocessing}), 
    which includes page segmentation, 
    layout element detection, 
    and OCR-based text recognition. 
Next, we perfrom Relation-Aware Table Annotation~(\ref{subsec:annotation_pipeline}), 
    where paragraphs associated with each table are manually annotated.
Based on the annotated data,
    we develop the Table-Text Association Model (TTAM), which captures semantic relationships between tables and textual content through learned semantic-level feature representations~(\ref{subsec:model_training}). 
Built upon TTAM and the structural preprocessing, 
    \sysname enables two downstream capabilities: 
    Table-Centric Semantic Parsing~(\ref{subsec:semantic_parsing}) and Domain-Specific Table Retrieval~(\ref{subsec:table_retrieval}).

\subsection{Document Structure Preprocessing}
\label{subsec:document_preprocessing}

PDF files are stored in binary format and typically lack structured semantic annotations, 
    making direct content analysis challenging~\cite{appalaraju2021docformer,huang2023improving}. 
To overcome this, 
    we adopt a mainstream image-based processing strategy~\cite{nassar2022tableformer}, 
    rendering each PDF page into an image to facilitate downstream structural and content analysis. 

Our pipeline begins by segmenting
    each PDF into individual pages and converting each page into an image. 
We then perform document layout analysis using an object detection 
    model fine-tuned on the \texttt{PubLayNet} dataset~\cite{zhong2019publaynet}, which segments each page into a set of visual blocks and classifies them into semantic categories.
\texttt{PubLayNet} follows an object detection annotation scheme, 
    labeling blocks as one of five classes: \textit{Text}, \textit{List}, \textit{Table}, \textit{Title}, or \textit{Figure}.
We use Faster R-CNN~\cite{ren2015faster} as the backbone detection model and fine-tune it on \texttt{PubLayNet} to enhance layout parsing accuracy. For all detected blocks classified as \textit{Text}, \textit{Title}, \textit{List}, or \textit{Table}, we apply Tesseract OCR~\cite{tesseract} to extract the textual content.
The resulting text and structural annotations serve as the foundation for modeling semantic associations between tables and relevant textual segments.

\subsection{Relation-Aware Table Annotation Pipeline}
\label{subsec:annotation_pipeline}

\vspace{0.5mm}
\noindent
\textbf{Relation-Aware Annotation.}
We annotate the parsed blocks produced during the Document Structure 
    Preprocessing stage, which includes page segmentation, block type classification, and OCR-based text extraction.
    Focusing on tables, we treat blocks labeled as \textit{Table} as anchors and manually identify their semantically associated textual descriptions.
Since \textit{List} blocks often contain descriptive content relevant 
    to tables, we annotate them alongside \textit{Text} blocks when evaluating their relationship to \textit{Table} blocks.
Throughout this paper, the term \textit{Text block} refers collectively to both \textit{Text} and \textit{List} blocks.

The annotation process follows these guidelines:

\begin{itemize}
\item \textbf{Number Matching:} Label any text block that explicitly references a table by its number as related;
\item \textbf{Semantic Supplement:} Include additional paragraphs that do not explicitly mention the table number but are semantically relevant;
\item \textbf{Completeness Check:} Ensure each table block has at least one associated text block. If none, determine whether it reflects an annotation oversight or a genuine lack of textual reference.
\end{itemize}

Each table annotation within a PDF document is represented as a triplet:
    <\texttt{Table-ID}, \texttt{Page-ID}, (\texttt{Related Paragraphs})>,
where \texttt{Table-ID} uniquely identifies a   
    \texttt{Table} block, \texttt{Page-ID} denotes the page that the table appears, and \texttt{(Related Paragraphs)} is the set of associated \textit{Text} and \textit{List} blocks.

To ensure annotation quality and reliability, we engaged 
    two researchers with over ten years of experience in document authoring and structured content analysis.
    They independently annotated table--text relationships to minimize potential errors and subjective bias stemming from limited domain knowledge or interpretation variance.

\vspace{0.5mm}
\noindent
\textbf{Annotation Results Reconciliation.}
To further ensure the reliability of the annotation results, 
    we adopted an expert consensus resolution strategy~\cite{pustejovsky2012natural} to reconcile discrepancies between annotators. 
Following the initial phase of independent annotation, the 
    two experts collaboratively reviewed all instances with conflicting labels.
Through in-depth discussion and mutual examination of their annotation rationales, 
    they reached a consensus on each disputed case to produce a finalized, 
    high-quality annotation set.

\subsection{Table-Text Association Model Training}
\label{subsec:model_training}

\vspace{0.5mm}
\noindent
\textbf{TTAM Model Structure.} The Table-Text Association Model (TTAM) takes as input a parsed and OCR-processed \textit{Table} block paired with a \textit{Text} block, 
    and outputs a binary classification indicating whether the text semantically describes the corresponding table.

Due to the inherent complexity of document semantics, 
    characterized by multiple semantic layers, 
    diverse information carriers, 
    and flexible referencing styles, 
    capturing such semantics poses significant challenges for existing shallow table recognition models and general-purpose natural language understanding systems.
These models often fall short in capturing document-level semantic structures, 
    particularly in the absence of datasets explicitly designed for this purpose.
To address this gap, 
    TTAM leverages an annotated table-text association dataset and builds upon pretrained natural language understanding models. This design enables TTAM to acquire table-centric structural and semantic knowledge, facilitating deep and context-aware semantic parsing within complex documents.

Specifically, 
    TTAM frames the table-text relation task as a sentence-pair classification problem, 
    where each input pair \( (b_{\mathrm{table}}, b_{\mathrm{text}}) \) consists of a table block and a text block. 
A pretrained model \( \mathcal{M} \) encodes this pair into a contextual representation, 
    which is passed to a classifier \( \mathcal{C} \) to predict whether the pair is ``related'' or ``unrelated'',
    as formalized in Equation~(\ref{eq:binary_classification}).
When combined with \sysname's document preprocessing module, 
    TTAM enables the parsing of inter-block semantic relationships, thereby facilitating document-level semantic understanding.
Notably, 
    TTAM is designed to be model-agnostic and has been successfully instantiated with various pretrained architectures, including BERT~\cite{devlin2019bert}, BART~\cite{lewis2020bart}, and RoBERTa~\cite{liu2019roberta}.
Its modular design allows for the integration of other transformer-based models, 
    offering adaptability to different analytic requirements and computational environments.

\begin{equation}
\hat{y} = \mathbb{I}\left\{\mathrm{Softmax}\left(\mathcal{C}\big(\mathcal{M}(b_{\mathrm{table}}, b_{\mathrm{text}})\big)\right) \geq \theta \right\}
\label{eq:binary_classification}
\end{equation}

\vspace{0.5mm}
\noindent
\textbf{Training Strategy.}
During training,
    we construct training samples based on the annotation results obtained from the Relation-Aware Table Annotation phase and train TTAM using a cross-entropy loss function.
Specifically,
    for each table identified by its \texttt{Table-ID}, 
    we create positive samples by pairing the corresponding \textit{Table} block with each associated \textit{Text} block from the annotated set of \texttt{(Related Paragraphs)}, indicating a semantic association.
To construct negative samples, 
    we randomly select an equal number of \textit{Text} blocks from the same document that are not included in the \texttt{(Related Paragraphs)} and pair them with the table block to denote non-association. 
This sampling strategy ensures a balanced distribution of positive and negative examples for effective training.

\begin{equation}
\mathcal{L}_{\text{CE}}(i) = -\big( y_i \log p_i + (1 - y_i) \log (1 - p_i) \big)
\label{eq:cross_entropy}
\end{equation}

Each constructed sample,
    consisting of a table block and a text block, 
    is treated as a sentence pair \( (b_{\mathrm{table}}, b_{\mathrm{text}}) \) and fed into the TTAM model for relation classification. 
Specifically,
    the sentence pair is first encoded by a pretrained language model \( \mathcal{M} \),
    which produces contextualized embeddings. These representations are then passed to a classifier \( \mathcal{C} \) to predict the probability \( p_i \) that the pair is the semantically ``related'' class. 
The model is optimized using the binary cross-entropy loss, 
    as defined in Equation~(\ref{eq:cross_entropy}), 
    where \( y_i \in \{0, 1\} \) denotes the ground-truth label of sample \( i \), and \( p_i \) is the predicted probability of the ``related'' class. 
This training process guides TTAM to effectively capture semantic associations between tables and text blocks, 
    enabling robust document-level semantic parsing.

\subsection{Table-Centric Semantic Parsing}
\label{subsec:semantic_parsing}

Building upon TTAM and the document structure preprocessing pipeline, 
    \sysname enables \textbf{Table-Centric Semantic Parsing},
    with the workflow depicted in Equation~(\ref{eq:semantic_parsing}).
Given a PDF document $D$, 
    \sysname first applies the preprocessing pipeline $\mathcal{P}(\cdot)$, which segments $D$ into discrete layout blocks and assigns semantic types,
    resulting in a block set $\mathcal{B} = \mathcal{P}(D)$. 
From $\mathcal{B}$, all blocks labeled as \textit{Table} are extracted as $\mathcal{T} = \{ b \in \mathcal{B} \mid \mathrm{type}(b) = \mathrm{Table} \}$. Each table block $t \in \mathcal{T}$ is treated as an anchor.
For each anchor $t$, 
    \sysname invokes TTAM to determine which \textit{Text} or \textit{List} blocks ---
    collectively denoted as $\mathcal{S} = \{ b \in \mathcal{B} \mid \mathrm{type}(b) \in \{\mathrm{Text}, \mathrm{List}\} \}$ --- that are semantically associated with $t$. 
The subset of related text blocks for table $t$ is then computed as $\mathcal{R}_t = \{ s \in \mathcal{S} \mid \mathrm{TTAM}(t, s) = 1 \}$. Through this process,
    \sysname performs fine-grained, table-centered semantic parsing, extracting each table along with its associated text, enabling comprehensive document-level semantic analysis.

{
\begin{equation}
\begin{aligned}
\mathrm{Parse}(D) = \{ (t, \mathcal{R}_t) \mid \; & t \in \mathcal{T}, \\
& \mathcal{R}_t = \{ s \in \mathcal{S} \mid \mathrm{TTAM}(t, s) = 1 \} \}
\end{aligned}
\label{eq:semantic_parsing}
\end{equation}
}

As illustrated in Figure~\ref{subfig:instance_a},
    this application example helps to demonstrate the full process of \sysname performing Table-Centric Semantic Parsing. 
The input PDF consists of multiple pages, 
    each containing various page elements. 
\sysname first conducts preprocessing,
    including page segmentation and layout analysis, 
    to detect individual page blocks and their corresponding types --- that is, 
    it successfully identifies the table within the PDF page as a table block.
It then uses each \textit{Table} block as an anchor to identify semantically associated \textit{Text} blocks across the document.

By analyzing the surrounding textual content, 
    \sysname successfully identifies six paragraphs that describe or interpret the content of the table block, 
    as highlighted in the figure.
In the era of large-scale data, semantic-level document parsing 
    offers a powerful approach for extracting salient information from complex, multi-modal documents, substantially reducing the manual effort required for downstream analysis and decision-making.

\subsection{Domain-Specific Table Retrieval}
\label{subsec:table_retrieval}

Another core capability of \sysname is \textbf{Domain-Specific Table Retrieval},
    which accepts as input a user-defined natural-language query and a target PDF document and returns as output a set of tables - 
    together with their associated descriptive text segments - 
    that are semantically relevant to the query.

To enable this functionality, 
\sysname first performs Table-Centric Semantic Parsing to segment the document into a set of candidate table blocks $\{ t_i \}_{i=1}^{N}$, 
each representing a distinct \textit{Table} region extracted through document layout analysis.
For semantic matching, 
\sysname adopts a fine-tuned RoBERTa cross-encoder to jointly encode the natural language query $q$ and each candidate table $t_i$. 
Specifically, each input pair $(q, t_i)$ is tokenized and fed into the encoder to produce a contextualized representation of the [CLS] token, 
which is further passed through a scoring layer to compute a scalar relevance score $s_i$, as formalized in Equation~(\ref{eq:ranking_score}):
\begin{align}
s_i &= \mathrm{Score}(q, t_i) \notag \\
    &= \mathbf{w}^{\top} \cdot \texttt{RoBERTa}_{\mathrm{CLS}}(q, t_i) + b, \quad \forall i = 1, \dots, N
\label{eq:ranking_score}
\end{align}

All candidate tables are then ranked based on their scores $\{s_i\}$ in descending order, 
and the top-$k$ tables are returned:

\begin{equation}
\mathcal{R}_{\mathrm{top}k} = \mathrm{TopK} \left( \left\{ (t_i, s_i) \right\}_{i=1}^{N} \right)
\label{eq:retrieval_ranking}
\end{equation}

As illustrated in Figure~\ref{subfig:instance_b},
the annual report of listed companies contains multiple tables, each presenting distinct information.
\sysname first segments the document into structured blocks and identifies all regions classified as \textit{Table}.
Each table block is paired with the user-defined query and encoded jointly using the cross-encoder.
The retrieval score is then computed via the scoring head, and top-ranked tables are returned.
This retrieval mechanism is trained using a margin-based ranking loss over positive and negative query--table pairs, 
ensuring that relevant tables receive higher scores than irrelevant ones.
Manual validation confirms that the top-ranked tables are consistently aligned with the query intent,
demonstrating the effectiveness of the semantic ranking framework.

In the era of large-scale, unstructured document corpora, this table-level retrieval capability offers an efficient and scalable solution for content navigation, 
alleviating the cognitive and computational burden for domain experts and analysts.

\section{Experiments}

\subsection{Implementation Details}
\vspace{0.05mm}
\noindent
\textbf{Document Structure Preprocessing.}
We employ pdf2image~\cite{pdf2image} to segment the document into individual pages and export them as \texttt{.jpg} images. 
In the document layout parsing stage, 
    we utilize Faster R-CNN to perform layout analysis on page images. 
The model is implemented within the Detectron2~\cite{detectron2} framework, 
    using the officially released \texttt{PubLayNet} dataset~\cite{publaynet_dataset}.

\vspace{0.05mm}
\noindent
\textbf{TTAM Implementation.}
TTAM leverages encoder-based pretrained models to extract feature representations from input data. 
Specifically, 
    it integrates three pretrained models, BERT~\cite{devlin2019bert} (\texttt{bert-based-uncased}),
    BART~\cite{liu2019roberta} (\texttt{bart-base}),
    and RoBERTa~\cite{lewis2020bart} (\texttt{roberta-base}).
The model downloading,
    deployment, 
and related operations are all implemented using the Hugging Face Transformers library~\cite{transformers}.

\vspace{0.05mm}
\noindent
\textbf{Experimental Environment.}
All experiments, 
    including model training and evaluation, 
    were conducted on a Ubuntu 22.04 server equipped with an RTX 4090 GPU.

\subsection{Experimental Settings}

\vspace{0.05mm}
\noindent
\textbf{Dataset.}
As no publicly available dataset currently exists for document-level semantic structure analysis,
    particularly with a focus on table-centric semantics,
    we constructed, 
    to the best of our knowledge, 
    the first dataset explicitly designed to model document semantic structures with tables as primary anchors.
This dataset was developed following the data annotation 
    pipeline detailed in Section~\ref{subsec:annotation_pipeline}. Specifically, we collected documents from the following two domains:

\begin{itemize}
    \item \textbf{arXiv}~\cite{arxiv}: An open-access repository of scholarly papers covering the natural sciences, engineering, and related fields. Specifically, in April 2025, we retrieved the 5,000 most recently uploaded paper PDFs from arXiv and randomly selected 130 of them, excluding those that employed uncommon formatting templates, as the subjects of our study.
    \item \textbf{PubMed Central}~\cite{pubmed_central}: An open-access database of literature in the life sciences and medical domains, offering a rich source of standardized, table-intensive documents. In April 2025, we retrieved the 5,000 most recently uploaded paper PDFs from PubMed Central and randomly selected 120 of them, again excluding those with non-standard formatting templates.
\end{itemize}

Table~\ref{table:dataset} summarizes the dataset statistics.  
\texttt{\#PDF}, \texttt{\#Page}, \texttt{\#Table Block}, and \texttt{\#Text Block} denote the number of source PDFs, total pages, extracted tables, and associated descriptive text blocks.

\vspace{0.5mm}
\noindent
\textbf{For TTAM Evaluation:}  
We annotated 3,248 table-text pairs (1,624 positive, 1,624 negative), and randomly split them (7:3) into 2,273 training and 975 test samples.

\vspace{0.5mm}
\noindent
\textbf{For Domain-Specific Table Retrieval:}  
From 100 sampled tables, two domain-specific queries were created per table -- one from the table title and one via expert consensus -- yielding 200 <query, table> pairs.  
After filtering incomplete or ambiguous cases, the final set includes 129 training and 53 test samples.

\begin{table}[t]
\centering
\caption{Details of the Constructed Dataset}
\begin{tabular}{c c c c c}
\toprule[1.5pt]
\textbf{Source} & \textbf{\#PDF} & \textbf{\#Page} & \textbf{\#Table Block} & \textbf{\#Text Block}  \\ \midrule
arXiv          &   125   &   2,408   &   741    &   1,101  \\
PubMed Central &   102   &    1,544      &   320       &   523       \\
Sum            &   227      &    3,952       &    1,061      &   1,624       \\
\bottomrule[1.5pt]
\end{tabular}
\label{table:dataset}
\end{table}

\vspace{0.05mm}
\noindent
\textbf{Baselines.}
As there is currently no established method in the academic 
    literature that analyzes the semantic structure of PDF documents using table-centric cues, we employ capable LLMs as experimental baselines.
Specifically, 
    we utilize GPT-4o, Gemini-2.0 Flash, and Claude 3.5, paired with a carefully constructed prompt to form our baseline evaluation framework.

\begin{itemize}
    \item \textbf{GPT-4o~\cite{gpt4o}:} 
    Developed by OpenAI, GPT-4o is a state-of-the-art multimodal model supporting text, vision, and audio inputs. Its strong understanding of tables and document layouts makes it a suitable baseline for this task.
    \item \textbf{Gemini-2.0 Flash~\cite{gemini}:} Proposed by Google DeepMind, this is a highly efficient multimodal model optimized for fast, high-quality processing of text and structured visual data, making it a strong candidate for baseline comparison.
    \item \textbf{Claude 3.5~\cite{claude}:} A multimodal language model capable of interpreting complex document structures, including tables, and is included as a baseline to assess semantic understanding in document parsing.
\end{itemize}

To enable the LLM to analyze the relationship between table blocks and text blocks,
    we designed the following prompt to guide the model’s understanding of the task and fully leverage its capabilities,
    in which \texttt{[table\_content]} and \texttt{[text\_content]} denote the OCR-scan results of table blocks and text blocks, respectively:

\begin{mdframed}[
    backgroundcolor=gray!20,
    linecolor=black,        
    linewidth=1pt,          
    innerleftmargin=10pt,   
    innerrightmargin=10pt,  
    innertopmargin=10pt,    
    innerbottommargin=10pt  
]
\textbf{\textit{Prompt:}} 
You are an expert in document analysis. Your task is to determine whether the provided text block is a descriptive explanation of the given table block.
Please reply with only a single number:

\noindent
Reply `1' if the text block describes or explains the table block.

\noindent
Reply `0' if the text block is unrelated to the table block.

\noindent
Here is the content:

- Table Block:
  [table\_content]

- Text Block:
  [text\_content]
\end{mdframed}

\vspace{0.05mm}
\noindent
\textbf{Metrics.}
We define the following metrics to quantitatively evaluate the performance of \sysname:

\begin{itemize}
    \item \textbf{Precision, Recall, and F1 of Text-Table Relation (\%):} Evaluate the TTAM's ability to correctly link text blocks to table blocks. Positive samples represent true associations, while negative samples represent unrelated pairs. Metrics are computed based on true positives (TP), false positives (FP), and false negatives (FN).
    \item \textbf{Document-Level Semantic Parsing Correctness:}  
    The number of PDF documents where table-text associations are correctly recognized, covering completely correctness and partly correctness.

    \item \textbf{Retrieval Recall@K (\%):}  
    Measures the proportion of relevant tables correctly retrieved within the top-K results, reflecting the effectiveness of the retrieval strategy.

    \item \textbf{Latency (s):} Measures the time overhead (in seconds) required for \sysname to complete the analysis.
\end{itemize}

\vspace{0.05mm}
\noindent
\textbf{Research Questions.}
To evaluate the performance of \sysname and compare it against baseline methods, 
    we define the following research questions \textbf{(RQs)} focusing on its TTAM model and two core functionalities: 
    Table-Centric Semantic Parsing and Domain-Specific Table Retrieval:

\begin{itemize}
    \item \textbf{RQ1:} Can \sysname's TTAM model effectively determine whether a text block describes a specific table block?
    \item \textbf{RQ2:} Can \sysname accurately perform semantic parsing of PDF documents using tables as structural cues?
    \item \textbf{RQ3:} Can \sysname reliably retrieve relevant tables and their contextual text based on user-provided natural language queries?
    \item \textbf{RQ4:} Does \sysname outperform the baselines in time efficiency and maintain low latency?
\end{itemize}

\subsection{RQ1: TTAM Performance}

\begin{table}[t]
\centering
\caption{Performance Evaluation of Table--Text Block Linking}
\begin{tabular}{c c c c c c c c}
\toprule[1.5pt]
\textbf{Scheme} & \textbf{TP} & \textbf{FP} & \textbf{TN} & \textbf{FN} & \textbf{Precision} & \textbf{Recall} & \textbf{F1} \\ \midrule
GPT-4o        &   168     &   19    &  450   &  338    &   89.84   &  33.20    &  48.48  \\
Gemini-2.0        &    373       &     59       &   410  &   133   &  86.34    &   73.72   &  79.53  \\
Claude-3.5        &     316      &     31       &  438   &   190   &  91.07    &   62.45   &   74.09 \\
BERT        &      426     &   35         & 434    &  80    &  92.41    &  84.19    &  88.11  \\
BART        &     444       &    50       &   419  &   62   & 89.88    &   87.75   & 88.80 \\
\textbf{RoBERTa}        &  \textbf{455}       &   \textbf{50}        &   \textbf{419} &  \textbf{51}    &   \textbf{90.10}  &  \textbf{89.92}    & \textbf{90.01} \\
\bottomrule[1.5pt]
\end{tabular}
\label{table:block_linking}
\end{table}

The evaluation results of TTAM are summarized in Table~\ref{table:block_linking}. 
TTAM supports multiple pretrained models as encoders,
    currently including BERT, BART, and RoBERTa. 
Across all configurations, 
    TTAM consistently achieves over 85\% F1 score,
    with RoBERTa delivering the best performance achieving: Precision of 90.10\%, Recall of 89.92\%, and F1 score of 90.01\%, 
    demonstrating strong capability in accurately identifying table and text blocks.
Error analysis
    reveals that TTAM's failure cases primarily involve overly generic text descriptions that lack specific references to table elements such as headers or numerical data.
For example,
    in document \texttt{Doc-A}~\footnote{The document name is anonymized in accordance with platform policies.}, 
    the table presents a fluctuation in mean absolute error relative to a variable. However, because the table contains minimal text (primarily numbers) and the accompanying paragraph only describes trends without citing specific values, TTAM incorrectly classifies the pair as unrelated. Notably, such cases are also difficult to resolve even through manual inspection.
For comparison, 
    we evaluated three state-of-the-art LLMs: GPT-4o, Gemini-2.0 Flash, and Claude 3.5.
While these models achieve relatively high precision. For example, 
    Claude 3.5 attains 91.07\% Precision, 
    indicating reliable identification of relevant table-text pairs -- 
    they suffer from substantial false negatives. Claude 3.5, in particular, produces 190 false negatives, resulting in a Recall of only 62.45\%,
    reflecting significant omissions of table-associated segments.

\subsection{RQ2: Document-Level Semantic Parsing}

In this section, 
    we conduct a document-level analysis of the TTAM test set to compare the performance of different methods from a table-centric perspective. 
Each document in the test set contains multiple table-text pairs, which may be either descriptively related or unrelated. 
We evaluate performance using the following three criteria: 
    (1) the number of documents in which all table-text relationships are correctly identified (\textbf{All Correct});
    (2) the number of documents in which all descriptively related table-text pairs are correctly identified, i.e., positive samples (\textbf{POS Correct}); 
    and (3) the number of documents in which all unrelated table-text pairs are correctly identified, i.e., negative samples (\textbf{NEG Correct}). 
These metrics respectively assess each method's capacity for comprehensive semantic structure analysis, accurate identification of relevant content, and avoidance of false associations. The evaluation is conducted on the TTAM test set, which includes 193 documents.

Results are shown in Table~\ref{table:document_parsing}. 
Evidently, 
    \sysname achieves the highest performance across \texttt{All Correct} and \texttt{POS Correct}, outperforming all three decoder-based LLMs.
This demonstrates \sysname's superior ability to extract 
    semantically relevant content using tables as anchors and its a stronger document-level semantic understanding compared to state-of-the-art generative models.
It is important to note that although LLMs tend to adopt conservative decision strategies, they often produce fewer false positives but more false negatives. As a result, they show relatively better performance on the NEG Correct metric in this limited test set. However, since the primary goal of semantic parsing is to accurately identify related table–text associations, the modest performance of LLM-based approaches in this area highlights their limitations.

\begin{table}[t]
\centering
\caption{Results of Table-Centric Document Semantic Parsing}
\begin{tabular}{c c c c c}
\toprule[1.5pt]
\textbf{Scheme} & \textbf{All Correct} & \textbf{POS Correct} & \textbf{NEG Correct} & \textbf{\#Sum} \\ \midrule
GPT-4o          &     109    &    119      &  183   & \multirow{4}{*}{193} \\
Gemini-2.0      &    113      &   147    &  159     &  \\
Claude-3.5      &    114       &  137       &   170  &    \\
\textbf{\sysname}        & \textbf{128}        & \textbf{166}          & \textbf{155}     &             \\
\bottomrule[1.5pt]
\end{tabular}
\label{table:document_parsing}
\end{table}

\begin{figure*}[th]
    \centering
    \begin{subcaptionbox}{Mean Time Overhead\label{subfig:mean}}[0.45\linewidth]
        {\includegraphics[width=\linewidth]{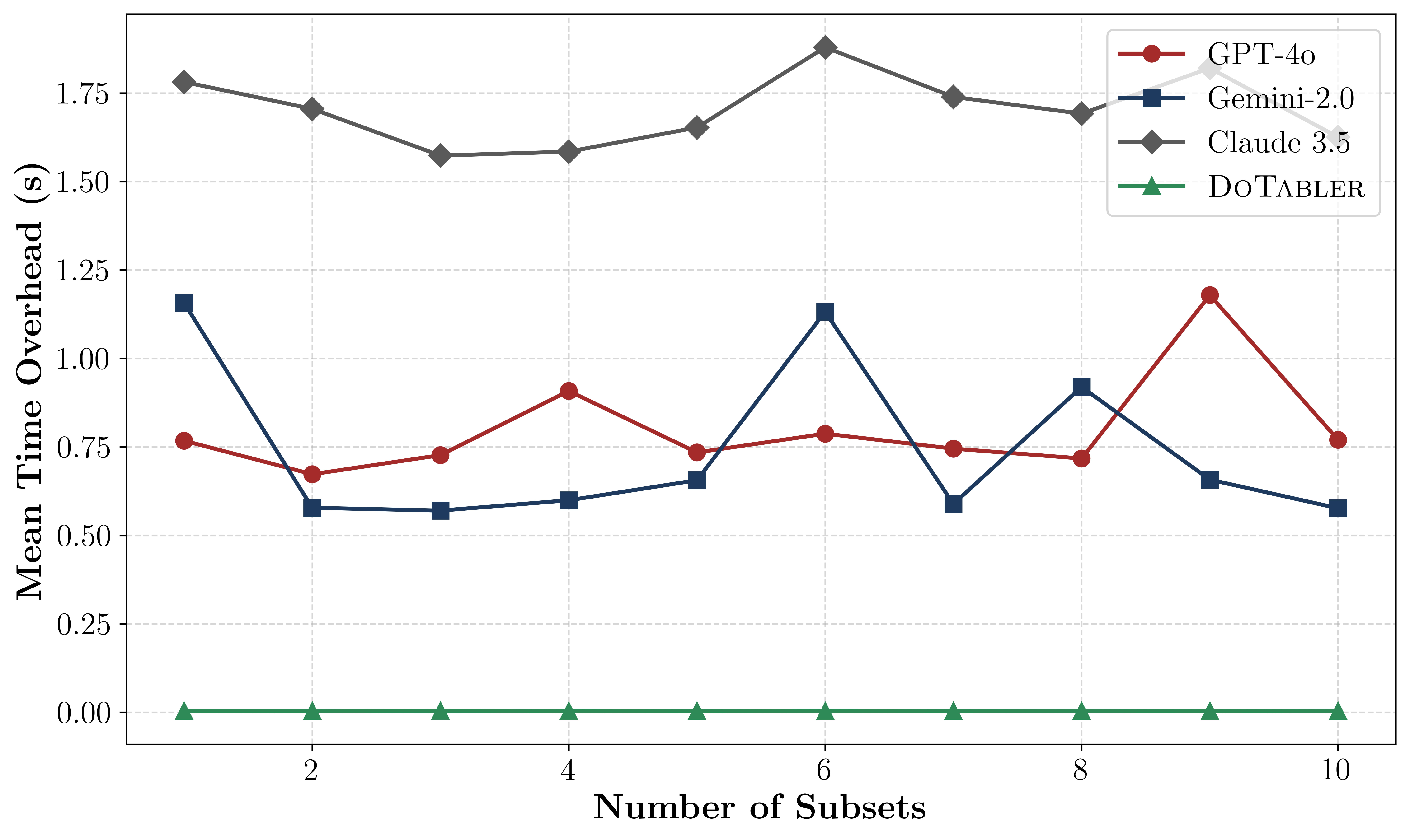}}
    \end{subcaptionbox}
    \hfill
    \begin{subcaptionbox}{Median Time Overhead\label{subfig:median}}[0.45\linewidth]
        {\includegraphics[width=\linewidth]{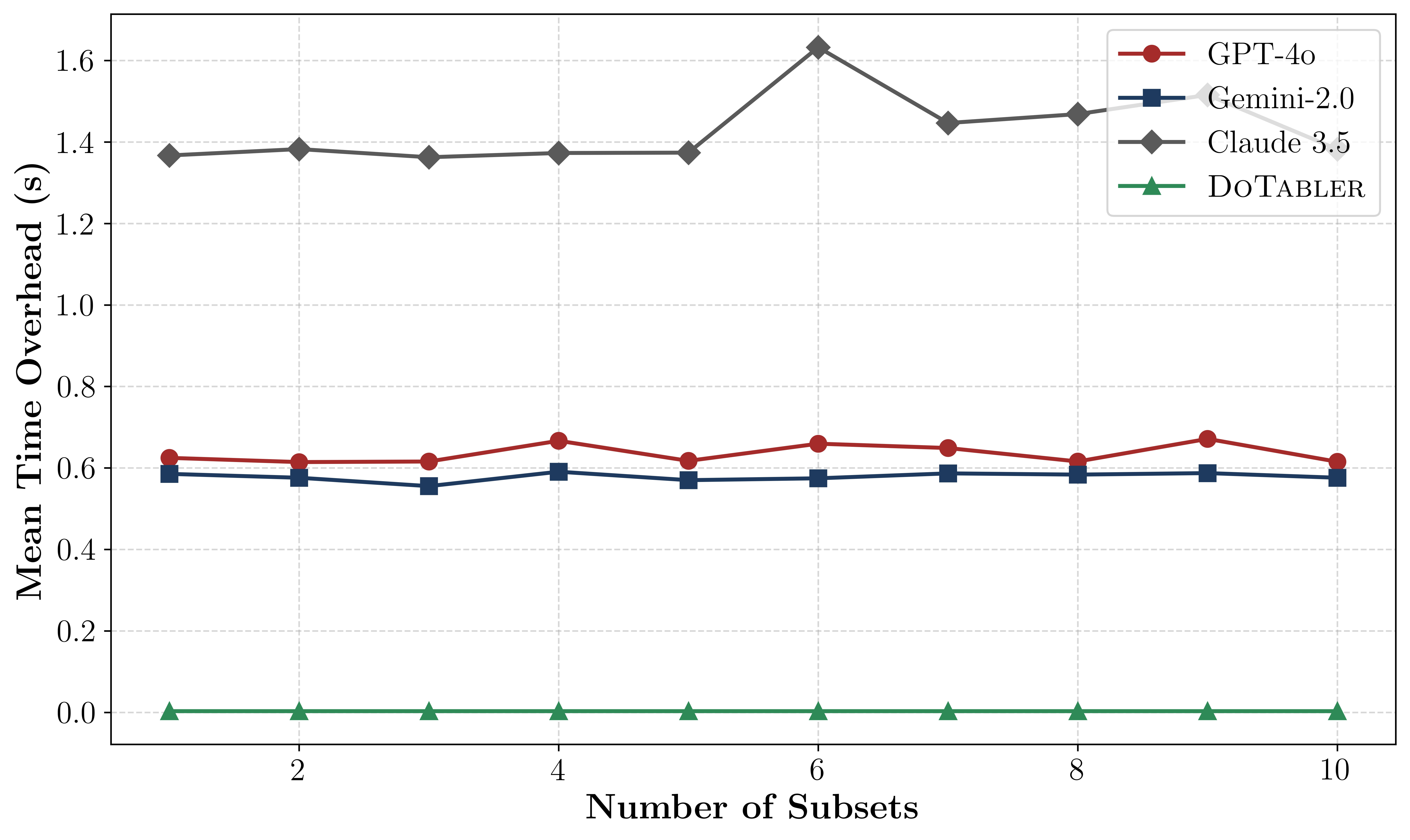}}
    \end{subcaptionbox}
    \vspace{10pt}
    \caption{Time Overhead of Distinct Schemes regarding Subsets}
    \label{fig:parsing_time_overhead}
    \vspace{10pt}
\end{figure*}

\subsection{RQ3: Domain-Specific Table Retrieval}

Table~\ref{table:table_retrieval} reports the results of the domain-specific table retrieval evaluation.
Given a natural language query, 
    \sysname employs a TTAM-based ranking strategy to compute the semantic relevance between the query and all tables within a PDF document, and returns the top-$K$ ranked tables as retrieval results.
When $K=1$ -- i.e.,
    retrieving only the most relevant table—the retrieval recall (Recall@1) reaches 71.70\%.
As $K$ increases to 3, the recall improves to 88.48\%.

It is worth noting that PDF documents often contain multiple structurally diverse tables; 
    in some extreme cases, 
    a single document may include a large number of tables.
For instance, 
    one test case contains 27 tables.
Accurately retrieving the query-relevant table under such conditions poses a significant challenge.
Nonetheless, 
    \sysname demonstrates strong robustness and practical effectiveness across these complex scenarios.

\begin{table}[t]
\centering
\caption{Results of Domain-Specific Table Retrieval}
\begin{tabular}{c c}
\toprule[1.5pt]
\textbf{Scheme} & \textbf{Retrieval Recall@K} \\ \midrule
\sysname@K=1    & 71.70                       \\
\sysname@K=2    & 84.91                       \\
\sysname@K=3    & 88.68                       \\
\bottomrule[1.5pt]
\end{tabular}
\label{table:table_retrieval}
\end{table}

\subsection{RQ4: Efficiency Evaluation}

In this section, 
    we evaluate the time overhead of \sysname on document semantic parsing and compare it with LLM-based approaches. Specifically,
    we measure both the average and median time overhead for each method using the 975 test pairs from RQ2 (semantic parsing). 
To mitigate the potential impact of data distribution bias, 
    we randomly divide all test samples into 10 batches evenly and compute the mean and median time overhead within each group.

The results are presented in Table~\ref{table:time_overhead}, Figure~\ref{subfig:mean} and Figure~\ref{subfig:median}. \sysname achieves both average and median time overheads below 0.01 seconds, demonstrating exceptional efficiency. 
This is largely attributed to TTAM, an encoder-based, moderately sized pre-trained model that runs locally during inference, allowing fast execution.
In contrast, 
    the three LLM-based baselines show significantly higher latency, with overheads approximately two orders of magnitude greater. For instance, Claude 3.5 exhibits average and median time overheads exceeding 1 second. In the batch-wise analysis shown in Figures~\ref{subfig:mean} and~\ref{subfig:median}, \sysname consistently outperforms all LLMs across both metrics, maintaining a significant time advantage. This result highlights the high efficiency of \sysname, which is particularly critical in real-world scenarios involving large volumes of documents with dense table–text structures.

\begin{table}[t]
\centering
\caption{Overall Time Cost of Distinct Schemes}
\begin{tabular}{ c c c}
\toprule[1.5pt]
\textbf{Scheme} & \textbf{Mean Time Cost (s)}         &\textbf{Median Time Cost (s)}               \\
                                \midrule
GPT-4o         &    0.8008   &   0.6201         \\
Gemini-2.0     &    0.7434   &   0.5763        \\
Claude-3.5     &    1.7054       &    1.4214          \\
\textbf{\sysname}       &    \textbf{0.0035}       &    \textbf{0.0031}     \\
\bottomrule[1.5pt]
\end{tabular}
\label{table:time_overhead}
\end{table}

\section{Discussion}

\subsection{Model Effectiveness and Comparative Analysis}

Through a series of comprehensive evaluations, we demonstrated the effectiveness of \sysname in capturing semantic associations between tables and related textual segments,
    highlighted the strong performance of \sysname in table-centric semantic parsing tasks.

While advanced generative language models such as GPT-4o and Gemini exhibit impressive reasoning and multimodal capabilities, \sysname -- powered by RoBERTa -- outperforms them in structured document understanding. Trained with masked language modeling, RoBERTa is well-suited for capturing fine-grained contextual dependencies and ensuring precise alignment between structured components. Its bidirectional encoder architecture and lower susceptibility to hallucinations allow it to excel in tasks like associating tables with relevant paragraphs.

In contrast, decoder-only models such as GPT-4o and Gemini are optimized for fluent text generation, which makes them more prone to hallucination -- producing outputs that are semantically plausible but factually inaccurate or unsupported~\cite{ji2023survey}. This weakness poses challenges in tasks requiring high-precision, cross-structural reasoning.
The superior performance of RoBERTa's discriminative approach has also been corroborated by other recent studies~\cite{benkler2024recognizing, roccabruna2024will, cheng2025beyond}, further validating its effectiveness in structured document analysis.

\subsection{Limitations and Future Works}

Despite its overall effectiveness, \sysname has certain limitations. First, it depends on existing preprocessing tools such as document layout analysis and OCR. While these techniques are generally reliable, they still struggle with complex layouts, non-standard templates, and scanned documents containing embedded tables. These challenges can affect parsing accuracy. Nonetheless, \sysname remains effective on the majority of documents evaluated.
Second, the performance of  \sysname may degrade on low-quality documents, especially when the relationship between tables and text is vague or implicit. In rare cases, contextual descriptions refer to general trends without explicitly mentioning table headers or values, making semantic association difficult. Future efforts may focus on improving robustness to irregular document structures and enhancing the model's ability to infer implicit semantic links.

\subsection{Ethical Considerations}

All data were sourced from publicly available arXiv and PubMed Central (Open Access Subset) documents, using only metadata and annotations in compliance with open-access licenses (e.g., CC-BY). No sensitive or personal information was included, and all data were used solely for academic research.

\section{Conclusion}
In this paper, we proposed \sysname, a table-centric 
    semantic document parsing framework that integrates multiple shallow-level document analysis modules, including document segmentation, layout analysis, and OCR. This is implemented through a three-stage pipeline centered around the TTAM. To evaluate the effectiveness of our approach, we constructed a dataset comprising nearly 4,000 pages of real-world PDF documents containing 1,000 tables. Experimental results show that \sysname achieves highly competitive performance, even when compared with advanced LLMs. As a general-purpose framework for table-centric semantic document parsing, \sysname has demonstrated strong capability in extracting tables and their associated textual context. In future work, we aim to enhance \sysname's capabilities and broaden its applicability across diverse domains and real-world deployment scenarios.

\clearpage
\bibliography{reference}
\end{document}